\documentclass[a4paper]{article}

\usepackage{INTERSPEECH2019}
\usepackage{tikz}
\usepackage{multirow}

\title{LipReading with 3D-2D-CNN BLSTM-HMM and word-CTC models}
\name{\begin{tabular}{c} Dilip Kumar Margam, Rohith Aralikatti, Tanay Sharma, Abhinav Thanda, \\ Pujitha A K, Sharad Roy, Shankar M Venkatesan\end{tabular}}
\address{Samsung R\&D Institute India, Bangalore}
\email{\{dilip.margam,r.aralikatti,abhinav.t89,tanay.sharma, \\ \hspace{1cm} pujitha.k,sharad.roy,s.venkatesan\}@samsung.com}

\begin{document}

\maketitle
\begin{abstract}
In recent years, deep learning based machine lipreading has gained prominence. To this end, several architectures such as LipNet, LCANet and others have been proposed which perform extremely well compared to traditional lipreading DNN-HMM hybrid systems trained on DCT features. 
In this work, we propose a simpler architecture of 3D-2D-CNN-BLSTM network with a bottleneck layer. 
We also present analysis of two different approaches for lipreading on this architecture. In the first approach, 3D-2D-CNN-BLSTM network is trained with CTC loss on characters (ch-CTC). 
Then BLSTM-HMM model is trained on bottleneck lip features (extracted from 3D-2D-CNN-BLSTM ch-CTC network) in a traditional ASR training pipeline.
In the second approach, same 3D-2D-CNN-BLSTM network is trained with CTC loss on word labels (w-CTC). The first approach shows that bottleneck features perform better compared to DCT features.
Using the second approach on Grid corpus' seen speaker test set, we report $1.3\%$ WER -- a $55\%$ improvement relative to LCANet. On unseen speaker test set we report $8.6\%$ WER which is $24.5\%$ improvement relative to LipNet. 
We also verify the method on a second dataset of $81$ speakers which we collected.
Finally, we also discuss the effect of feature duplication on BLSTM-HMM model performance.
\end{abstract}
\noindent\textbf{Index Terms}: Bidirectional Long Short-Term Memory (BLSTM), Connectionist Temporal Classification (CTC), ASR, HMM, CNN

\section{Introduction}
\label{sec:intro}

The process of using only visual information of lip movements to convert speech to text is called machine lipreading.
The visual information such as lip movements, facial expression, tongue and teeth movements helps us in understanding other person's speech when audio is interrupted or corrupted.
Thus visual information acts as complimentary information to speech signal and helps in improving speech recognition systems in noisy environment \cite{fifteen,thirteen,aralikatti2018global}.
Machine lipreading has several practical applications such as silent movie transcription, aids for hearing-impaired persons, speech recognition in noisy environment.

For such reasons, machine lipreading has been a subject of research for last few decades \cite{five} and has gained greater prominence with the advent of deep learning.

One of the fundamental challenges in lipreading is to distinguish the words which are visually similar (called \textit{homophemes}) such as pack, back and mac (they sound different but have similar lip movements).
Such visual similarity acts as a hurdle for human lipreaders resulting in as low as $20\%$ accuracy \cite{eleven}.
On the other hand, recent advances in deep learning has shown remarkable improvements in performance of machine lipreading \cite{three,seven,fourteen}.

In this work we present a simpler lipreading architecture with comparable/better performance over traditional and existing state-of-the-art methods. The architecture contains 3D-CNN, 2D-CNN, and bi-directional LSTM layers where one 2D-CNN layer acts as a bottleneck layer.
We analyze two different approaches involving this architecture. 

In the first approach, we first train 3D-2D-CNN-BLSTM architecture in an end-to-end fashion with CTC loss\cite{twentytwo}  on character labels using curriculum learning. Similar to \cite{sixteen} we extract bottleneck features from the network and train a BLSTM-HMM hybrid model using these features.
The motivation behind this approach is to compare end-to-end CTC based training with traditional BLSTM-HMM hybrid model trained using cross-entropy loss on HMM labels with the same bottleneck features.

Recent works\cite{audhkhasi2017direct,soltau2016neural} on CTC based end-to-end ASR models trained with word labels instead of characters/phonemes have demonstrated promising improvements in WERs. Following this, in our second approach we train the 3D-2D-CNN-BLSTM network in an end-to-end fashion with CTC loss but using word labels instead of character labels used by \cite{seven,fourteen}. The reasoning behind this method is that, the drop in WERs due to homophemes can be mitigated by using whole words instead of characters (as contextual information can disambiguate homophemes). Accordingly, we observe significant improvements using words as output labels.

In addition to the above two approaches, we also discuss the importance of feature duplication in BLSTM-HMM hybrid model. Feature duplication was discussed in \cite{fifteen,sixteen,huang2013audio} but in the context of audio-visual feature fusion. This was done to match the visual frame-rate to the acoustic frame-rate. However, in our work we show that even for pure lipreading feature duplication can be useful.

The paper is organized as follows: In section \ref{sec:relwork}, we present the related work.
Section \ref{sec:models} explains the architecture of 3D-2D-CNN-BLSTM network. 
The first approach and second approach are presented in subsections \ref{sec:rnnhmm} and \ref{sec:wordctc} respectively.
The explanation about datasets and experiments is given in section \ref{sec:experiments}.
Section \ref{sec:results} presents the results and analysis. Section \ref{sec:conclusion} gives the conclusion.

\section{Related Work}
\label{sec:relwork}
 
Two decades ago, lipreading is seen as a word classification problem, where each input video is classified to one of the limited words.
The authors in \cite{one} do word classification using different variations of 3D CNN architectures. Word classification using CNNs followed by RNNs or HMMs is presented in number of different papers \cite{five,ten,twenty}. 
Later same authors proposed \textit{Watch, Listen, Attend and Spell (WLAS)} network in \cite{three}, which uses encoder-decoder type of architecture for audio-visual sentence level speech recognition. They have also introduced curriculum learning \cite{twentyone}, a strategy to accelerate training and reduce overfitting. We have adopted curriculum learning from this paper, which has resulted in faster convergence.

The end-to-end sentence-level lipreading with 3D-CNN-RNN based model (LipNet) with CTC loss on character labels is proposed in paper \cite{seven}. We also propose end-to-end sentence-level lipreading with a new 3D-2D-CNN-BLSTM network architecture in this paper which has fewer parameters compared to lipnet. We also train our model with CTC loss on word labels. 
%Different approaches using DCT and AAM visual features in traditional speech-style GMM-HMM models for audio-visual speech recognition are discussed in \cite{nine}. We use 3D-2D-CNN-BLSTM network bootle-neck features to train RNN-HMM models.
The paper \cite{nine} presents different experiments which use DCT and AAM visual features in traditional speech-style GMM-HMM models for audio-visual speech recognition. But in this paper we use 3D-2D-CNN-BLSTM network features instead of DCT or AMM visual features in RNN-HMM context.
Several papers  \cite{twelve} have tried phoneme or viseme labels for CTC loss, followed by WFST with language model. Two commonly used cost functions in end-to-end based sequential models are CTC-loss and sequence-to-sequence loss. Comparison between CTC-loss and sequence-to-sequence loss in different conditions is shown in \cite{thirteen} using 3D-CNN and LSTM based architectures. The conditional independence assumption in CTC loss is said to be one of cons for CTC based sequential models. 
The LCANet proposed in \cite{fourteen} has used highway network with bidirectional GRU layers after 3D CNN layers with cascaded attention-CTC to overcome conditional independence assumption of CTC.

The CTC loss with word labels is explored for ASR task in \cite{seventeen,audhkhasi2017direct,soltau2016neural} but was not attempted for lipreading. In this paper we attempt lipreading with CTC loss on word labels and discuss its limitations. In \cite{fifteen} we used a technique of feature duplication for DNN-HMM model with DCT features of lip image, to match the frame rate of audio signal. But we haven't explored the significance of feature duplication in HMM based models. In this paper we show how feature duplication in context of HMM based models helps in improving performance. In another paper \cite{sixteen} we used CTC loss on character labels with DCT features of lip as input to RNN layers.

\section{Models}
\label{sec:models}

\subsection{3D-2D-CNN-BLSTM Network}
\label{sec:3d2dnetwork}

% Below is an example of how to insert images. Delete the ``\vspace'' line,
% uncomment the preceding line ``\centerline...'' and replace ``imageX.ps''
% with a suitable PostScript file name.
% -------------------------------------------------------------------------
\begin{figure*}[htb]

\begin{minipage}[b]{1.0\linewidth}
  \centering
  \begin{tikzpicture}[scale=0.6, transform shape]
\def\x{-5}\def\y{-0.5}\def\h{1.5}
\def\w{2}\def\d{4}\def\b{2.85}
\coordinate (A) at (\x,\y);
\coordinate (B) at (\x,\y+\h);
\coordinate (C) at (\x+\w,\y+\h+\d);
\coordinate (D) at (\x+\w,\y+\d);

\coordinate (A1) at (\x+\b,\y);
\coordinate (B1) at (\x+\b,\y+\h);
\coordinate (C1) at (\x+\w+\b,\y+\h+\d);
\coordinate (D1) at (\x+\w+\b,\y+\d);

\coordinate (B2) at (\x-0.2+\w/4,\y+0.2+\h+\d/4);
\coordinate (C2) at (\x-0.2+\w*3/4,\y+0.2+\h+\d*3/4);

\draw[dashed] (A) -- (D);
\node[anchor=south west, inner sep=0] (lip2) at (D) {\includegraphics[scale=0.8]{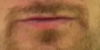}};
\node[anchor=south west, inner sep=0] (lip1) at (\x+3*\w/4,\y+3*\d/4) {\includegraphics[scale=0.8]{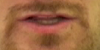}};
\node[anchor=south west, inner sep=0] (lip1) at (\x+2*\w/4,\y+2*\d/4) {\includegraphics[scale=0.8]{lip02}};
\node[anchor=south west, inner sep=0] (lip1) at (\x+\w/4,\y+\d/4) {\includegraphics[scale=0.8]{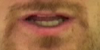}};
\node[anchor=south west, inner sep=0] (lip0) at (A) {\includegraphics[scale=0.8]{lip00}};
\draw[dashed] (B) -- (C);
\draw[dashed] (B1) -- (C1);
\draw[dashed] (A1) -- (D1);
\draw[white] (A) -- node[below,black] {Input video} (A1);
\draw[->,black!70,thick] (B2) -- node[above,sloped] {time} (C2);

\draw [->,>	=stealth,thick,black!70] (-0.7,2.2) -- (0.3,2.2);

%3DCNN1 layers
\def\x{0}\def\y{-0.5}\def\h{2}
\def\w{2}\def\d{4}\def\b{2}
\coordinate (A) at (\x,\y);
\coordinate (B) at (\x,\y+\h);
\coordinate (C) at (\x+\w,\y+\h+\d);
\coordinate (D) at (\x+\w,\y+\d);

\coordinate (A1) at (\x+\b,\y);
\coordinate (B1) at (\x+\b,\y+\h);
\coordinate (C1) at (\x+\w+\b,\y+\h+\d);
\coordinate (D1) at (\x+\w+\b,\y+\d);

%\fill [draw=none, fill=gray!50] (\x+\w,\y+\d) -- (\x+\w,\y+\h+\d) -- (\x+\w+\b,\y+\h+\d) -- (\x+\w+\b,\y+\d) -- cycle;
\def\num{16}
\foreach \i in {8,...,1}
{
\fill [draw=none, fill=gray!20] (\x+2*\i*\w/\num,\y+2*\i*\d/\num) -- (\x+2*\i*\w/\num,\y+\h+2*\i*\d/\num) -- (\x+2*\i*\w/\num+\b,\y+\h+2*\i*\d/\num) -- (\x+2*\i*\w/\num+\b,\y+2*\i*\d/\num) -- cycle;
\fill [draw=none, fill=gray!80] (\x+2*\i*\w/\num - \w/\num,\y+2*\i*\d/\num - \d/\num) -- (\x+2*\i*\w/\num - \w/\num,\y+\h+2*\i*\d/\num - \d/\num) -- (\x+2*\i*\w/\num - \w/\num+\b,\y+\h+2*\i*\d/\num - \d/\num) -- (\x+2*\i*\w/\num - \w/\num+\b,\y+2*\i*\d/\num - \d/\num) -- cycle;
}

\fill [draw=none, fill=gray!20] (A) -- (B) -- (B1) -- (A1) -- cycle; 
%\fill [draw=none, fill=white] (B) -- (C) -- (C1) -- (B1) -- cycle;
%\fill [draw=none, fill=white] (A1) -- (B1) -- (C1) -- (D1) -- cycle;
\draw[dashed,gray] (A) -- (B) -- (C);
\draw[dashed,gray] (A) -- node[below,black] {3D-Conv1} (A1);
\draw[dashed,gray] (B) -- (B1);
\draw[dashed,gray] (C) -- (C1);
\draw[dashed,gray] (A1) -- (B1) -- (C1) -- (D1) -- cycle;

%3DPool1 layers
\def\x{3}\def\y{0}\def\h{1.2}
\def\w{2}\def\d{4}\def\b{1.2}
\coordinate (A) at (\x,\y);
\coordinate (B) at (\x,\y+\h);
\coordinate (C) at (\x+\w,\y+\h+\d);
\coordinate (D) at (\x+\w,\y+\d);

\coordinate (A1) at (\x+\b,\y);
\coordinate (B1) at (\x+\b,\y+\h);
\coordinate (C1) at (\x+\w+\b,\y+\h+\d);
\coordinate (D1) at (\x+\w+\b,\y+\d);

\def\num{10}
\foreach \i in {5,...,1}
{
\fill [draw=none, fill=gray!30] (\x+2*\i*\w/\num,\y+2*\i*\d/\num) -- (\x+2*\i*\w/\num,\y+\h+2*\i*\d/\num) -- (\x+2*\i*\w/\num+\b,\y+\h+2*\i*\d/\num) -- (\x+2*\i*\w/\num+\b,\y+2*\i*\d/\num) -- cycle;
\fill [draw=none, fill=gray!90] (\x+2*\i*\w/\num - \w/\num,\y+2*\i*\d/\num - \d/\num) -- (\x+2*\i*\w/\num - \w/\num,\y+\h+2*\i*\d/\num - \d/\num) -- (\x+2*\i*\w/\num - \w/\num+\b,\y+\h+2*\i*\d/\num - \d/\num) -- (\x+2*\i*\w/\num - \w/\num+\b,\y+2*\i*\d/\num - \d/\num) -- cycle;
}

\fill [draw=none, fill=gray!20] (A) -- (B) -- (B1) -- (A1) -- cycle; 
%\fill [draw=none, fill=white] (B) -- (C) -- (C1) -- (B1) -- cycle;
%\fill [draw=none, fill=white] (A1) -- (B1) -- (C1) -- (D1) -- cycle;
\draw[dashed,gray] (A) -- (B) -- (C);
\draw[dashed,gray] (A) -- node[below,black] {3D-Pool1} (A1);
\draw[dashed,gray] (B) -- (B1);
\draw[dashed,gray] (C) -- (C1);
\draw[dashed,gray] (A1) -- (B1) -- (C1) -- (D1) -- cycle;

\draw [->,>	=stealth,thick,black!70] (5.5,2.3) -- (6.3,2.3);

%3DCNN2 layers
\def\x{6}\def\y{0}\def\h{1.2}
\def\w{2}\def\d{4}\def\b{1.2}
\coordinate (A) at (\x,\y);
\coordinate (B) at (\x,\y+\h);
\coordinate (C) at (\x+\w,\y+\h+\d);
\coordinate (D) at (\x+\w,\y+\d);

\coordinate (A1) at (\x+\b,\y);
\coordinate (B1) at (\x+\b,\y+\h);
\coordinate (C1) at (\x+\w+\b,\y+\h+\d);
\coordinate (D1) at (\x+\w+\b,\y+\d);

\def\num{16}
\foreach \i in {8,...,1}
{
\fill [draw=none, fill=gray!20] (\x+2*\i*\w/\num,\y+2*\i*\d/\num) -- (\x+2*\i*\w/\num,\y+\h+2*\i*\d/\num) -- (\x+2*\i*\w/\num+\b,\y+\h+2*\i*\d/\num) -- (\x+2*\i*\w/\num+\b,\y+2*\i*\d/\num) -- cycle;
\fill [draw=none, fill=gray!80] (\x+2*\i*\w/\num - \w/\num,\y+2*\i*\d/\num - \d/\num) -- (\x+2*\i*\w/\num - \w/\num,\y+\h+2*\i*\d/\num - \d/\num) -- (\x+2*\i*\w/\num - \w/\num+\b,\y+\h+2*\i*\d/\num - \d/\num) -- (\x+2*\i*\w/\num - \w/\num+\b,\y+2*\i*\d/\num - \d/\num) -- cycle;
}

\fill [draw=none, fill=gray!20] (A) -- (B) -- (B1) -- (A1) -- cycle; 
%\fill [draw=none, fill=white] (B) -- (C) -- (C1) -- (B1) -- cycle;
%\fill [draw=none, fill=white] (A1) -- (B1) -- (C1) -- (D1) -- cycle;
\draw[dashed,gray] (A) -- (B) -- (C);
\draw[dashed,gray] (A) -- node[below,black] {3D-Conv2} (A1);
\draw[dashed,gray] (B) -- (B1);
\draw[dashed,gray] (C) -- (C1);
\draw[dashed,gray] (A1) -- (B1) -- (C1) -- (D1) -- cycle;

%3DPool2 layers
\def\x{8}\def\y{0.3}\def\h{0.8}
\def\w{2}\def\d{4}\def\b{0.8}
\coordinate (A) at (\x,\y);
\coordinate (B) at (\x,\y+\h);
\coordinate (C) at (\x+\w,\y+\h+\d);
\coordinate (D) at (\x+\w,\y+\d);

\coordinate (A1) at (\x+\b,\y);
\coordinate (B1) at (\x+\b,\y+\h);
\coordinate (C1) at (\x+\w+\b,\y+\h+\d);
\coordinate (D1) at (\x+\w+\b,\y+\d);

\def\num{10}
\foreach \i in {5,...,1}
{
\fill [draw=none, fill=gray!30] (\x+2*\i*\w/\num,\y+2*\i*\d/\num) -- (\x+2*\i*\w/\num,\y+\h+2*\i*\d/\num) -- (\x+2*\i*\w/\num+\b,\y+\h+2*\i*\d/\num) -- (\x+2*\i*\w/\num+\b,\y+2*\i*\d/\num) -- cycle;
\fill [draw=none, fill=gray!90] (\x+2*\i*\w/\num - \w/\num,\y+2*\i*\d/\num - \d/\num) -- (\x+2*\i*\w/\num - \w/\num,\y+\h+2*\i*\d/\num - \d/\num) -- (\x+2*\i*\w/\num - \w/\num+\b,\y+\h+2*\i*\d/\num - \d/\num) -- (\x+2*\i*\w/\num - \w/\num+\b,\y+2*\i*\d/\num - \d/\num) -- cycle;
}

\fill [draw=none, fill=gray!20] (A) -- (B) -- (B1) -- (A1) -- cycle; 
%\fill [draw=none, fill=white] (B) -- (C) -- (C1) -- (B1) -- cycle;
%\fill [draw=none, fill=white] (A1) -- (B1) -- (C1) -- (D1) -- cycle;
\draw[dashed,gray] (A) -- (B) -- (C);
\draw[dashed,gray] (A) -- node[below,black] {3D-Pool2} (A1);
\draw[dashed,gray] (B) -- (B1);
\draw[dashed,gray] (C) -- (C1);
\draw[dashed,gray] (A1) -- (B1) -- (C1) -- (D1) -- cycle;

\draw [->,>	=stealth,thick,black!70] (10,2.4) -- (11,2.4);

%2DConv1 layers
\def\x{10.5}\def\y{0.5}\def\h{0.6}
\def\w{2}\def\d{4}\def\b{0.6}
\coordinate (A) at (\x,\y);
\coordinate (B) at (\x,\y+\h);
\coordinate (C) at (\x+\w,\y+\h+\d);
\coordinate (D) at (\x+\w,\y+\d);

\coordinate (A1) at (\x+\b,\y);
\coordinate (B1) at (\x+\b,\y+\h);
\coordinate (C1) at (\x+\w+\b,\y+\h+\d);
\coordinate (D1) at (\x+\w+\b,\y+\d);

\def\num{22}
\foreach \i in {11,...,1}
{
\fill [draw=none, fill=gray!20] (\x+2*\i*\w/\num,\y+2*\i*\d/\num) -- (\x+2*\i*\w/\num,\y+\h+2*\i*\d/\num) -- (\x+2*\i*\w/\num+\b,\y+\h+2*\i*\d/\num) -- (\x+2*\i*\w/\num+\b,\y+2*\i*\d/\num) -- cycle;
\fill [draw=none, fill=gray!80] (\x+2*\i*\w/\num - \w/\num,\y+2*\i*\d/\num - \d/\num) -- (\x+2*\i*\w/\num - \w/\num,\y+\h+2*\i*\d/\num - \d/\num) -- (\x+2*\i*\w/\num - \w/\num+\b,\y+\h+2*\i*\d/\num - \d/\num) -- (\x+2*\i*\w/\num - \w/\num+\b,\y+2*\i*\d/\num - \d/\num) -- cycle;
}

\fill [draw=none, fill=gray!20] (A) -- (B) -- (B1) -- (A1) -- cycle; 
%\fill [draw=none, fill=white] (B) -- (C) -- (C1) -- (B1) -- cycle;
%\fill [draw=none, fill=white] (A1) -- (B1) -- (C1) -- (D1) -- cycle;
\draw[dashed,gray] (A) -- (B) -- (C);
\draw[dashed,gray] (A) -- node[below,black] {2D-Conv1} (A1);
\draw[dashed,gray] (B) -- (B1);
\draw[dashed,gray] (C) -- (C1);
\draw[dashed,gray] (A1) -- (B1) -- (C1) -- (D1) -- cycle;

%2DConv2 layers
\def\x{12}\def\y{0.6}\def\h{0.4}
\def\w{2}\def\d{4}\def\b{0.4}
\coordinate (A) at (\x,\y);
\coordinate (B) at (\x,\y+\h);
\coordinate (C) at (\x+\w,\y+\h+\d);
\coordinate (D) at (\x+\w,\y+\d);

\coordinate (A1) at (\x+\b,\y);
\coordinate (B1) at (\x+\b,\y+\h);
\coordinate (C1) at (\x+\w+\b,\y+\h+\d);
\coordinate (D1) at (\x+\w+\b,\y+\d);

\def\num{22}
\foreach \i in {11,...,1}
{
\fill [draw=none, fill=gray!20] (\x+2*\i*\w/\num,\y+2*\i*\d/\num) -- (\x+2*\i*\w/\num,\y+\h+2*\i*\d/\num) -- (\x+2*\i*\w/\num+\b,\y+\h+2*\i*\d/\num) -- (\x+2*\i*\w/\num+\b,\y+2*\i*\d/\num) -- cycle;
\fill [draw=none, fill=gray!80] (\x+2*\i*\w/\num - \w/\num,\y+2*\i*\d/\num - \d/\num) -- (\x+2*\i*\w/\num - \w/\num,\y+\h+2*\i*\d/\num - \d/\num) -- (\x+2*\i*\w/\num - \w/\num+\b,\y+\h+2*\i*\d/\num - \d/\num) -- (\x+2*\i*\w/\num - \w/\num+\b,\y+2*\i*\d/\num - \d/\num) -- cycle;
}

\fill [draw=none, fill=gray!20] (A) -- (B) -- (B1) -- (A1) -- cycle; 
%\fill [draw=none, fill=white] (B) -- (C) -- (C1) -- (B1) -- cycle;
%\fill [draw=none, fill=white] (A1) -- (B1) -- (C1) -- (D1) -- cycle;
\draw[dashed,gray] (A) -- (B) -- (C);
\draw[dashed,gray] (A) -- node[below,black] {2D-Conv2} (A1);
\draw[dashed,gray] (B) -- (B1);
\draw[dashed,gray] (C) -- (C1);
\draw[dashed,gray] (A1) -- (B1) -- (C1) -- (D1) -- cycle;

%\node[draw] at (\x+\b/2,\y-0.5) {bottleneck features};
\node[align=left] at (14.1,2) {bottleneck\\ features};

\draw [->,>	=stealth,thick,black!70] (13.5,2.5) -- node[above,black] {linearize} (15,2.5);

%BLSTM1 layers
\def\x{14.5}\def\y{0.3}\def\h{1}
\def\w{2}\def\d{4}\def\b{0.3}
\coordinate (A) at (\x,\y);
\coordinate (B) at (\x,\y+\h);
\coordinate (C) at (\x+\w,\y+\h+\d);
\coordinate (D) at (\x+\w,\y+\d);

\coordinate (A1) at (\x+\b,\y);
\coordinate (B1) at (\x+\b,\y+\h);
\coordinate (C1) at (\x+\w+\b,\y+\h+\d);
\coordinate (D1) at (\x+\w+\b,\y+\d);

\coordinate (B2) at (\x+\w/5+\b,\y+\h/2+\d/5);
\coordinate (C2) at (\x+\w*4/5+\b,\y+\h/2+\d*4/5);

%\fill [draw=none, fill=white] (A) -- (B) -- (B1) -- (A1) -- cycle; 
%\fill [draw=none, fill=white] (B) -- (C) -- (C1) -- (B1) -- cycle;
%\fill [draw=none, fill=white] (A1) -- (B1) -- (C1) -- (D1) -- cycle;
\draw[black!70] (A) -- (B) -- (C);
\draw[black!70] (A) -- node[below,black] {BLSTM1} (A1);
\draw[black!70] (B) -- (B1);
\draw[black!70] (C) -- (C1);
\draw[black!70] (A1) -- (B1) -- (C1) -- (D1) -- cycle;
%\draw[->,thick] (B2) -- (C2);

%BLSTM2 layers
\def\x{16}\def\y{0.3}\def\h{1}
\def\w{2}\def\d{4}\def\b{0.3}
\coordinate (A) at (\x,\y);
\coordinate (B) at (\x,\y+\h);
\coordinate (C) at (\x+\w,\y+\h+\d);
\coordinate (D) at (\x+\w,\y+\d);

\coordinate (A1) at (\x+\b,\y);
\coordinate (B1) at (\x+\b,\y+\h);
\coordinate (C1) at (\x+\w+\b,\y+\h+\d);
\coordinate (D1) at (\x+\w+\b,\y+\d);

\coordinate (B2) at (\x+\w/5+\b,\y+\h/2+\d/5);
\coordinate (C2) at (\x+\w*4/5+\b,\y+\h/2+\d*4/5);

\fill [draw=none, fill=white] (A) -- (B) -- (B1) -- (A1) -- cycle; 
\fill [draw=none, fill=white] (B) -- (C) -- (C1) -- (B1) -- cycle;
\fill [draw=none, fill=white] (A1) -- (B1) -- (C1) -- (D1) -- cycle;
\draw[black!70] (A) -- (B) -- (C);
\draw[black!70] (A) -- node[below,black] {BLSTM2} (A1);
\draw[black!70] (B) -- (B1);
\draw[black!70] (C) -- (C1);
\draw[black!70] (A1) -- (B1) -- (C1) -- (D1) -- cycle;
%\draw[->,thick] (B2) -- (C2);

\draw [->,>	=stealth,thick,black!70] (17.5,2.5) -- (18.5,2.5);

%linear/softmax layer
\def\x{18}\def\y{0.3}\def\h{1}
\def\w{2}\def\d{4}\def\b{0.3}
\coordinate (A) at (\x,\y);
\coordinate (B) at (\x,\y+\h);
\coordinate (C) at (\x+\w,\y+\h+\d);
\coordinate (D) at (\x+\w,\y+\d);

\coordinate (A1) at (\x+\b,\y);
\coordinate (B1) at (\x+\b,\y+\h);
\coordinate (C1) at (\x+\w+\b,\y+\h+\d);
\coordinate (D1) at (\x+\w+\b,\y+\d);

\fill [draw=none, fill=white] (A) -- (B) -- (B1) -- (A1) -- cycle; 
\fill [draw=none, fill=white] (B) -- (C) -- (C1) -- (B1) -- cycle;
\fill [draw=none, fill=white] (A1) -- (B1) -- (C1) -- (D1) -- cycle;
\draw (A) -- (B) -- (C);
\draw (A) -- node[align=center, below, black] {linear \\ with softmax} (A1);
\draw (B) -- (B1);
\draw (C) -- (C1);
\draw (A1) -- (B1) -- (C1) -- (D1) -- cycle;

%CTC Loss
%\def\x{21.5}\def\y{0.75}\def\h{1}
%\def\w{2}\def\d{4}\def\b{0.3}
%\coordinate (A) at (\x,\y);
%\coordinate (B) at (\x,\y+\h);
%\coordinate (C) at (\x+\w,\y+\h+\d);
%\coordinate (D) at (\x+\w,\y+\d);
%
%\coordinate (A1) at (\x+\b,\y);
%\coordinate (B1) at (\x+\b,\y+\h);
%\coordinate (C1) at (\x+\w+\b,\y+\h+\d);
%\coordinate (D1) at (\x+\w+\b,\y+\d);
%
%
%\fill [draw=none, fill=white] (A) -- (B) -- (B1) -- (A1) -- cycle; 
%\fill [draw=none, fill=white] (B) -- (C) -- (C1) -- (B1) -- cycle;
%\fill [draw=none, fill=white] (A1) -- (B1) -- (C1) -- (D1) -- cycle;
%\draw (A) -- (B) -- (C);
%\draw (A) -- node[align=center, below, black] {CTC \\ loss} (A1);
%\draw (B) -- (B1);
%\draw (C) -- (C1);
%\draw (A1) -- (B1) -- (C1) -- (D1) -- cycle;

  \end{tikzpicture}
  %\centerline{\includegraphics[width=17cm]{3D2Darchitecture.png}}
%  \vspace{2.0cm}
%  \centerline{(a) Result 1}\medskip
\end{minipage}
\caption{Architecture of 3D-2D-CNN-BLSTM network}
\label{fig:3d2dnetwork}
\end{figure*}

In this section we describe the architecture of proposed 3D-2D-CNN-BLSTM network. 
This network contains two 3D CNN layers, two 2D CNN layers followed by two BLSTM layers as shown in the Figure \ref{fig:3d2dnetwork}. In 3D convolution kernel moves along time, height and width dimensions of input. Whereas in 2D convolution kernel only moves along height and width dimensions.
%We propose new 3D-2D network architecture, which has two 3D CNN layers followed by two 2D CNN layers followed by two bidirectional LSTM layers as shown in the figure \ref{fig:3d2dnetwork}.
We apply batchnorm layer on input directly instead of applying any mean or variance normalization as in \cite{seven}, so that batchnorm layer can learn best suitable normalization parameters for the dataset. The detailed architecture is presented in Table \ref{tab:3d2dnetwork}.

\begin{table}[htp]
\centering
\caption{Detailed architecture of 3D-2D-CNN-BLSTM Network. N: number of frames in video, L: number of labels}
\label{tab:3d2dnetwork}
{\renewcommand{\arraystretch}{1.2}
\begin{tabular}{|c|c|c|} 
\hline
\textbf{Layer} & \textbf{Output size} & \textbf{Kernel/Stride/Pad} \\\hline
\hline
Input & Nx100x50x3 &  \\
batchnorm & Nx100x50x3 &  \\
\hline
3D-Conv1 & Nx50x25x32 & 3x5x5/1,2,2/1,2,2  \\
batchnorm/relu & Nx50x25x32 &  \\
3D-Pool1 & Nx25x12x32 & 1x2x2/1,2,2 \\
\hline
3D-Conv2 & Nx25x12x64 & 4x5x5/1,1,1/1,2,2  \\
batchnorm/relu & Nx25x12x64 &  \\
3D-Pool2 & Nx12x6x64 & 1x2x2/1,2,2 \\
\hline 
2D-Conv1 & Nx6x3x128 & 5x5/2,2/2,2 \\
batchnorm/relu & Nx6x3x128 & \\
\hline
2D-Conv2 & Nx3x2x8 & 3x3/2,2/2,2 \\
batchnorm/relu & Nx3x2x8 & \\
\hline
BLSTM1 & Nx400 &  \\
BLSTM2 & Nx400 &  \\
\hline
Linear & NxL &  \\
Softmax & NxL &  \\
\hline
%\bottomrule
\end{tabular}
}
\end{table}

The 3D-2D-CNN-BLSTM network has spatiotemporal convolution in fist two layers to consider contextual information in temporal dimension. Every 3D CNN and 2D CNN layer is followed by batchnorm layer. We don't use any pooling layer after 2D CNN layer. There is 3D maxpool layer after every 3D convolutional layer. The output of 2D-Conv2 is linearized and fed to two bidirectional LSTM layers. Softmax activation is applied on final CTC labels to get probabilities. The cell size for LSTM layers is $200$. The hyperbolic tangent activation function (Tanh) is used in LSTM layers.

The network is trained end-to-end with CTC loss. The input of the network is a sequence of $100\times50\times3$ dimensional RGB lip images. In the first approach CTC loss is computed on sequence of character labels and in second approach we compute CTC loss on sequence of word labels.

This network is used for obtaining bottleneck lip-features from 2D-Conv2 layer (described in Table \ref{tab:3d2dnetwork}) for BLSTM-HMM training.
The bottleneck features are of $48$ ($= 3 \times 2 \times 8$) dimensions obtained by linearizing outputs of 2D-Conv2. The lower dimensional lip-features will enable us to train a better BLSTM-HMM hybrid model.
%speech-style GMM-HMMs to get HMM state labels. The BLSTM-HMM is finally trained on these HMM state labels.

%\subsection{CTC loss}
%CTC is a loss function that eliminates the need for alignments between input sequence and the output sequence . 
%The CTC \cite{twentytwo} computes the probability of a sequence by marginalising over all sequences that are defined as equivalent to this sequence.
%The CTC contains the additional blank label $\phi$ along with the required output labels.
%For example all the ouput sequences of the form $(a,\phi,b,\phi,c)$, $(a,\phi,\phi,b,c)$, $(a,a,b,\phi,c)$ and $(a,b,\phi,c,\phi)$ maps to $(a,b,c)$.

\subsubsection{Curriculum Learning}
Curriculum learning has been shown to provide better performance and faster convergence\cite{twentyone}. In curriculum learning, the strategy is to train the network with segmented words for initial epochs, then in later epochs the network is trained on full sentences.
In our approach, each segmented word in the training set is used for training in a given epoch $eph$ with a probability $\max(0,1-eph/20)$. Thus with each epoch the expected number of segmented words reduce linearly.  
First epoch (epoch $0$) includes only segmented words. And from second epoch all full sentences are included. Curriculum learning has helped us in faster convergence, for 3D-2D-CNN-BLSTM w-CTC (described in subsection \ref{sec:wordctc}) model with curriculum learning it took $45$ epochs to converge whereas without curriculum learning it took $89$ epochs to converge. Similarly for other experiments also convergence has become twice as faster by using curriculum learning, as faster as by not using curriculum learning..

\subsection{BLSTM-HMM Hybrid Model}
\label{sec:rnnhmm}
%The 3D-2D-CNN-BLSTM network trained with CTC loss on character labels is used for extracting lip features.
%The output of 2D-Conv2 layer in 3D-2D-CNN-BLSTM network is of dimension $48$ and these $48$ dimensional features are extracted for each frame in each video.
In this approach, the $48$ dimensional bottleneck features are extracted for each input frame from 3D-2D-CNN-BLSTM network trained with ch-CTC. 
Then each feature is duplicated four times. Thus input sequence of length $l$
is converted to a sequence of length $4\times l$. 
Following the standard hybrid model training pipeline, a GMM-HMM model is trained on these features for context-independent phones (mono-phone GMM-HMM model). Bootstraped by the alignments generated from mono-phone model tri-phone GMM-HMM (a model with context-dependent phones) is trained. Then GMM-HMM with LDA transformed features is trained.

\begin{figure}[hbt]

\begin{minipage}[b]{1.2\linewidth}
\centering
  \begin{tikzpicture}[scale=1]
\def\h{0.6}
\coordinate (A) at (0,0);
\coordinate (B) at (4,\h);
\coordinate (A1) at (0,\h);
\coordinate (B1) at (4,2.5*\h);
\coordinate (A2) at (0,2.5*\h);
\coordinate (B2) at (4,3.5*\h);
\coordinate (A3) at (0,3.5*\h);
\coordinate (B3) at (4,4.5*\h);
\coordinate (A4) at (0,4.5*\h);
\coordinate (B4) at (4,5.5*\h);
\coordinate (A5) at (0,5.5*\h);
\coordinate (B5) at (4,6.5*\h);
\draw[black!70,line width=1.2pt,rounded corners=6pt] (A) rectangle (B) node[align=center,pos=0.5,black] { \textbf{bottleneck features} };
\draw[black!70,line width=1.2pt,rounded corners=6pt] (A1) rectangle (B1) node[align=center,pos=0.5,black] { \textbf{LDA transformation and} \\ \textbf{append} $\boldsymbol{\Delta, \Delta \Delta}$ };
\draw[black!70,line width=1.2pt,rounded corners=6pt] (A2) rectangle (B2) node[align=center,pos=0.5,black] { \textbf{BLSTM} };
\draw[black!70,line width=1.2pt,rounded corners=6pt] (A3) rectangle (B3) node[align=center,pos=0.5,black] { \textbf{BLSTM} };
\draw[black!70,line width=1.2pt,rounded corners=6pt] (A4) rectangle (B4) node[align=center,pos=0.5,black] { \textbf{HCLG WFST decoder} };
\draw[black!70,line width=1.2pt,rounded corners=6pt] (A5) rectangle (B5) node[align=center,pos=0.5,black] { \textbf{text output} };
\draw[black!70,line width=1.2pt,->] (5,0.5*\h) -- (5,6*\h);  
  \end{tikzpicture}
\end{minipage}
\caption{Training pipeline of BLSTM-HMM model with bottleneck features obtained from 3D-2D-CNN-BLSTM as input}
\label{fig:rnnhmm}
\end{figure}
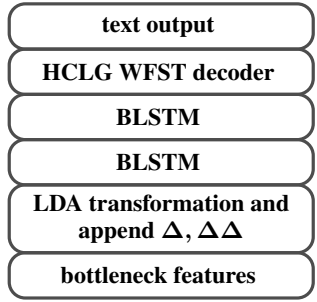

Finally, HMM tied-state labels obtained from GMM-HMM trained with LDA transforms are used for training BLSTM-HMM hybrid model with cross-entropy loss. The preprocessing involves applying cumulative mean-variance normalization and LDA producing $40$ dimensional features. Then $\Delta$, $\Delta \Delta$ features are appended to get $120$ dimensional features.
The BLSTM-HMM model is trained on these $120$d input features. The standard HCLG WFST is used for decoding. The complete pipeline is shown in Figure \ref{fig:rnnhmm}.

\subsubsection{Feature Duplication}
We observed that duplicating each input feature exactly $4$ times gives dramatic improvements for hybrid models. 
The Table \ref{tab:represults} shows how feature duplication results in nearly $50\%$ relative improvement for all methods. We believe that the duplicated frames help in disambiguating lip movements for each phoneme under different context. 

%(f_{1_1},f_{1_2},f_{1_3},f_{1_4},f_{2_5},f_{2_6},f_{2_7},f_{2_8},
%f_{3_9},f_{3_{10}},f_{3_{11}},f_{3_{12}})$
%For example consider a video with $3$ frames $(f_1,f_2,f_3)$. After duplication frame sequence looks like $(f_{1},f_{1},f_{1},f_{1},f_{2},f_{2},f_{2},f_{2},f_{3},f_{3},f_{3},f_{3})$ with $3\times4=12$ frames. 
%Then LDA is done on new set of features generated by appending $5$ contextual frames on both sides for every frame (concatenation of $5+1+5 = 11$ frames). LDA transform is used to reduce dimensionality of input features to $40$.
%Hence, the each frame after LDA dimensionality reduction is resulted from a combination of different $11$ frames.
%Hence, the resulted $6^{th}$ frame after LDA is a function of $11$ frames as show in equation \ref{equ:featdup}. 
%\begin{align}
%\label{equ:featdup}
%f_{new_6} &= \\  
%& LDA(f_{1_1},f_{1_2},f_{1_3},f_{1_4},f_{2_5},f_{2_6},f_{2_7},f_{2_8},
%f_{3_9},f_{3_{10}},f_{3_{11}})
%\end{align}
%Each new frame formed is different from other frame, resulting from different context.
%Then $\Delta$ and $\Delta \Delta$ are appended for each feature forming completely a new set of features. Finally, a sequnce of $12$ frames is created from sequence of $3$ frames. This is analogous to increasing ($4\times fr$) frame rate of a given video with frame rate $fr$. 
It is known that videos with higher frame rate give better performance for lipreading \cite{chictu2007influence}. Because of this, we believe BLSTM-HMM trained with feature duplication has shown significant improvement in performance. We intend to explore more on feature duplication in our futur work by duplicating it different number of times (like twice, thrice, etc,.) and analize it's effect on performance.

\begin{table}[htp]
\centering
\caption{Description of datasets used in our experiments}
\label{tab:dataset}
{\renewcommand{\arraystretch}{1.5}
\begin{tabular}{|c|c|c|c|} 
\hline
\textbf{Dataset} & \textbf{Vocabulary} & \textbf{\#utterances} & \textbf{\#speakers} \\\hline
\hline
\small{Grid} & 51 & 33000 & 33 \\\hline
\small{\shortstack{Indian \\ English}} & 236 & 14073 & 81 \\\hline
\end{tabular}
}
\end{table}

\subsection{Word-CTC Model}
\label{sec:wordctc}
The 3D-2D-CNN-BLSTM w-CTC (word-CTC) model is the 3D-2D-CNN-BLSTM model trained with CTC objective function on word labels. 
Each unique word in the given dataset is considered as different label. Additionally we have two more labels for space and blank.
For each input video frame, network associates a word label or a space or a blank label. 
The resulting output will be a sequence of words. 
The motivation for this approach is to overcome the conditional independence assumption of CTC loss.
The outputs of CTC are conditionally independent hence sometimes resulting in meaningless sequence of characters when character labels are used.

\section{Experiments}
\label{sec:experiments}
\subsection{Datasets}
\subsubsection{Grid}
The Grid audio-visual dataset is the widely used data for audio-visual or visual speech recognition tasks \cite{grid}.
Each sentence of Grid has a fixed structure with six words structured as: \textit{command + color + preposition + letter + digit + adverb}. For example \textit{"set red with m six please"}.
The dataset has $51$ unique words.
The $51$ words include four commands, four colors, four prepositions, 25 letters, ten digits, and four adverbs. Each sentence is randomly chosen combination of these words.
The duration of each utterance is $3$ seconds. The Table \ref{tab:dataset} gives more details about Grid corpus. 
For proper comparison, we employ same methods as of \cite{seven} for splitting train and test sets. 
For overlapped speakers (seen speakers), we randomly select $255$ utterances from each speaker to test set and remaining utterances are used for training.
For speaker independent case (unseen speakers), we held out two male speakers ($1$ and $2$) and two female speakers ($20$ and $22$) for evaluation and remaining speakers are used for training.
%Training and test datasets

\subsubsection{Indian English Dataset}
Indian English dataset is an audio-visual dataset we collected where $81$ speakers have spoken $180$ utterances each.
This dataset is collected by asking users to hold phone in their hand in comfortable manner, with only condition that their face should be visible in the video. Each speaker has to speak the utterance shown on the screen, video is recorded through front camera. 
Speakers are even allowed to move and talk. The utterances are choosen such that they are phonetically balanced. Hence Indian English Dataset (In-En) is more wilder dataset than Grid audio-visual dataset.
The length of utterance vary from $2-5$ seconds. Table \ref{tab:dataset} gives more details about the dataset.
For training we use $65$ speakers and for evaluation we use $16$ speakers out of which $4$ are female and $12$ are male speakers.
All the experiments on Indian English dataset presented in this paper are done on gray-scale videos with cumulative mean-variance normalization.

\subsection{Implementation Details}
For extracting the ROI (Region of Interest), we perform lip detection using YOLO v2 model \cite{yolo} trained to detect lip region. The ROI is then resized to size $100 \times 50$.

We implemented 3D-2D-CNN-BLSTM in Tensorflow \cite{tensorflow}. For 3D-2D-CNN-BLSTM network training we augment data with horizontally mirrored video frames. 
To make the network resilient to frame rate, we randomly drop or repeat the frames with probability $0.05$.
We use ADAM optimizer \cite{adam} with initial learning rate of $0.0001$ and batch size of $32$. We use native tensorflow CTC beam search decoder with beam width of $200$. For both ch-CTC model and w-CTC model same parameters are used. Before computing WER for ch-CTC model we correct spellings of words in predicted output by replacing it with closest word in dictionary using edit distance.

GMM-HMM models are trained using Kaldi toolkit \cite{kaldi}. Preprossesing steps involve LDA dimensionality reduction to $40$d from $11\times 48$d ($5$ contextual frames on both sides). Then $\Delta$ and $\Delta \Delta$ features are appended and BLSTM is trained on these features. 
The BLSTM in BLSTM-HMM is trained using Tensorflow. 
The ADAM optimizer with initial learning rate of $0.001$ and batch size $32$ is used. 
Two layer BLSTM is trained with cell dimension of 150. The outputs of BLSTM are posterior probabilities of HMM states.
The DCT BLSTM-HMM model is the baseline model where top $10\times10$ DCT features of $100\times50$ lip image are used as features in BLSTM-HMM model, same as in \cite{fifteen}.

\begin{table}[htp]
\centering
\caption{Performance on Grid and Indian English (In-En) datasets. \textsc{na}: \textit{results not available.}}
\label{tab:results}
{\renewcommand{\arraystretch}{1.8}
\begin{tabular}{|c|c|c|c|} 
\hline
\multirow{2}{*}{\textbf{Method}} & \multicolumn{2}{c|}{\textbf{\shortstack{Grid \\ (WER $\%$)}}} &   \textbf{\shortstack{In-En \\ (WER $\%$)}} \\
\cline{2-4}
 & \textbf{\small{seen}} & \textbf{\small{unseen}} & \textbf{\small{unseen}} \\
\hline \hline
\small{\shortstack{DCT \\ BLSTM-HMM}} & 8.9 & 16.5 & 20.8 \\\hline
\small{\shortstack{\textbf{3D-2D-CNN-BLSTM} \\ ch-CTC}} & 3.2 & 15.2 & 19.6 \\\hline
\small{\shortstack{\textbf{3D-2D-CNN} \\ \textbf{BLSTM-HMM}}} & 5.2 & 13.6 & 16.4 \\\hline
\small{LIPNET \cite{seven}} & 4.8 & 11.4 & \textsc{na} \\\hline
\small{WAS (WLAS) \cite{three}} & 3.0 & \textsc{na} & \textsc{na} \\\hline
\small{LCANet \cite{fourteen}} & 2.9 & \textsc{na} & \textsc{na} \\\hline
\small{\shortstack{\textbf{3D-2D-CNN-BLSTM} \\ \textbf{w-CTC}}} & \textbf{1.3} & \textbf{8.6} & \textbf{12.3} \\\hline
\end{tabular}
}
\end{table}

\section{Results}
\label{sec:results}
Results are reported on seen test set and on unseen test set for Grid dataset. For Indian English (In-En) dataset we report results on unseen dataset. The results for our proposed approaches compared with baseline DCT BLSTM-HMM and with LIPNET, LCANet, WAS (WLAS) are presented in Table \ref{tab:results}.

It is shown that 3D-2D-CNN-BLSTM w-CTC approach has achieved state-of-the-art performance with $55.1\%$ relative improvement over LCANet on Grid Dataset (on seen test set) and relative improvement of $24.5\%$ on unseen test set as compared to LIPNET. 
The 3D-2D-CNN BLSTM-HMM ch-CTC model has shown significant relative improvement over baseline DCT BLSTM-HMM model on Grid dataset.
The 3D-2D-CNN-BLSTM w-CTC is shown to perform relatively $25.0\%$ better than 3D-2D-CNN BLSTM-HMM model on Indian English dataset.
The Indian English dataset is wilder dataset where speakers are allowed to move and talk, unlike Grid. 
Our proposed approaches show similar trends in performance on both datasets, proving that our approaches are robust.
We have also tested BLSTM-HMM model with bottleneck features obtained from 3D-2D-CNN-BLSTM w-CTC model and results are very similar to the model trained on bottlenect features from 3D-2D-CNN-BLSTM ch-CTC.

\begin{table}[htp]
\centering
\caption{Effect of feature duplication on BLSTM-HMM model}
\label{tab:represults}
{\renewcommand{\arraystretch}{1.8}
\begin{tabular}{|c|c|c|c|} 
\hline
\textbf{Method} & \textbf{Dataset} & \textbf{\small{\shortstack{Without \\ Duplication}}} & \textbf{\small{\shortstack{With \\ Duplication}}} \\\hline
\hline
\small{\shortstack{DCT \\ BLSTM-HMM}} & Grid &25.5 & \textbf{8.9} \\\hline
\small{\shortstack{3D-2D-CNN \\ BLSTM-HMM}} & Grid & 8.6 & \textbf{5.3} \\\hline
\small{\shortstack{3D-2D-CNN \\ BLSTM-HMM}} & \shortstack{Indian \\ English} & 29.7 & \textbf{16.4} \\\hline
\end{tabular}
}
\end{table}

As discussed in subsection \ref{sec:rnnhmm} feature duplication in context of BLSTM-HMM models has shown relative improvement of $53\%$ and $44\%$ on Grid and Indian English datasets respectively as shown in Table \ref{tab:represults}.

Eventhough w-CTC model has given best performance, this approach is not scalable with vocabulary size and addition of new words to vocabulary is difficult. For such realword problems with vocabulary size of $1$ million ch-CTC or sub-word labels should be used. We intend to extend our work to sub-word label CTC loss in future and explore its potential.

\section{CONCLUSION}
\label{sec:conclusion}
We proposed new 3D-2D-CNN-BLSTM architecture, which is comparable to LCANet on Grid when ch-CTC loss is used.
The proposed 3D-2D-CNN-BLSTM w-CTC has given state-of-the-art results with relative improvement of $55\%$ and $24.5\%$ on Grid seen and unseen test sets with $1.3\%$ WER and $8.6\%$ WER respectively. 
%The motivation for trying word labels (w-CTC) is to avoid words with random spellings as output, which happens when character labels are used because of conditional independence assumption of CTC.
We also demonstrated that 3D-2D-CNN BLSTM-HMM models performs better than 3D-2D-CNN-BLSTM ch-CTC and DCT BLSTM-HMM models. 
The importance of duplicating features in context of HMM based models is also discussed. We have also mentioned the limitations of w-CTC models and possible alternatives.

\bibliographystyle{IEEEtran}

\bibliography{refs}

\end{document}